\documentclass{svproc}
\usepackage{rotating}
\usepackage{url}
\usepackage{algorithm}
\usepackage[noend]{algpseudocode}
\usepackage{tabularx}
\usepackage{amsmath}
\usepackage{graphicx}
\usepackage{booktabs}  
\usepackage{siunitx}   

\usepackage{amsmath,amssymb}

\begin{document}
\mainmatter              
\title{AGEL-Comp: A Neuro-Symbolic Framework \\ for Compositional Generalization in\\ Interactive Agents}
\titlerunning{AGEL-Comp: A Neuro-Symbolic Agent for Compositionality}  
%

\author{Mahnoor Shahid \and Hannes Rothe}
\authorrunning{Shahid and Rothe} 

\institute{Universität Duisburg-Essen, Essen, Germany \\
\email{\{mahnoor.shahid,hannes.rothe\}@uni-due.de}}

\maketitle              

\begin{abstract}
Large Language Model (LLM)-based agents exhibit systemic failures in compositional generalization, limiting their robustness in interactive environments. This work introduces AGEL-Comp, a neuro-symbolic AI agent architecture designed to address this challenge by grounding actions of the agent. AGEL-Comp integrates three core innovations: (1) a dynamic Causal Program Graph (CPG) as a world model, representing procedural and causal knowledge as a directed hypergraph; (2) an Inductive Logic Programming (ILP) engine that synthesizes new Horn clauses from experiential feedback, grounding symbolic knowledge through interaction; and (3) a hybrid reasoning core where an LLM proposes a set of candidate sub-goals that are verified for logical consistency by a Neural Theorem Prover (NTP). Together, these components operationalize a deduction--abduction learning cycle:  enabling the agent to deduce plans and abductively expand its symbolic world model, while a neural adaptation phase keeps its reasoning engine aligned with new knowledge.
We propose an evaluation protocol within the \texttt{Retro Quest} simulation environment to probe for compositional generalization scenarios to evaluate our AGEL agent. Our findings clearly indicate the better performance of our AGEL model over pure LLM-based models. Our framework presents a principled path toward agents that build an explicit, interpretable, and compositionally structured understanding of their world.

\end{abstract}


\section{Introduction}

Large Language Models (LLMs) have demonstrated remarkable capabilities across a wide range of natural language tasks~\cite{brown2020language,vaswani2017attention}. However, as these models are increasingly deployed as the cognitive core for interactive agents, a fundamental limitation becomes apparent: a failure in compositional generalization—the ability to understand and produce novel combinations from known, primitive components ~\cite{keysers2020measuring,sakai2025revisiting}. 
This ``compositionality crisis" is a critical barrier to achieving human-like intelligence, as it reveals a reliance on statistical pattern matching rather than a structured, systematic understanding of the world ~\cite{bender2020climbing,chomsky1965aspects,fodor1988connectionism}.  This brittleness arises because the models' knowledge is inherently disembodied and detached from the structured, causal world. Their understanding lacks empirical grounding, the connection between abstract symbols and their real-world referents, interactions, and consequences that govern how concepts systematically combine. 
Current models, including state-of-the-art LLMs, consistently struggle with these challenges, limiting their robustness and adaptability in open-ended, interactive environments~\cite{dziri2023faith,li2025morphological}.

To address this gap, we propose \textbf{AGEL-Comp} (Action-Grounded Experiential Learning for Compositionality), a novel neuro-symbolic framework for LLM-powered agents designed explicitly to foster compositional reasoning.
We argue that true generalization in interactive settings cannot be achieved through neural scaling alone but requires an architecture that can learn, represent, and reason over explicit, structured knowledge. By enabling an agent to learn from the consequences of its actions in an interactive environment, we can augment its pre-existing knowledge with an explicit world model that captures the compositional ``grammar" of its environment. 

AGEL-Comp integrates the flexibility of neural models with the rigor of symbolic systems through three core components: (1) a dynamic Causal Program Graph (CPG) as its world model $W$, representing procedural and causal knowledge in a modular, hierarchical structure conducive to complex planning~\cite{cabalar2014causal,wu2025causal,fujitsu2024causal}; (2) an Inductive Logic Programming (ILP) engine, $\mathcal{G}$, that synthesizes new, generalizable sub-programs (Horn clauses) from sparse experiential feedback, providing a formal mechanism for grounding symbolic knowledge through interaction~\cite{cropper2022inductive,muggleton1994inductive}; (3) a hybrid reasoning core where an LLM acts as a high-level planner, whose proposals are verified for logical consistency by a Neural Theorem Prover (NTP)~\cite{s_r6,s_r5,rocktaschel2017end}.

By building this model from grounded experience, the agent can move beyond shallow statistical pattern matching to systematic, rule-based reasoning. While prior work has focused on integrating LLMs into simulated environments for task completion, our primary contribution is the design and evaluation of a synergistic architecture that explicitly targets compositional generalization. 

We evaluate our framework within the \texttt{Retro Quest} simulation environment, using a new protocol specifically designed to probe compositional generalization with interactions. Our work makes the following novel contributions: 
\begin{enumerate}
    \item The AGEL-Comp architecture, a hybrid cognitive framework for LLM-based agents that enforces compositional reasoning.
    \item A hybrid deductive-abductive learning cycle that enables agents to learn from interaction by combining symbolic world model expansion with neural reasoner adaptation.
    \item A two-stage grounding mechanism that transforms raw experience into symbolic rules via causal attribution and inductive synthesis.
\end{enumerate}
Our results demonstrate that this neuro-symbolic approach significantly outperforms purely neural baselines, presenting a principled path toward agents that build an interpretable and compositionally structured understanding of their world.

\textbf{Paper overview.}
Section~2 reviews the three building blocks underlying AGEL-Comp (causal graphical modeling, inductive logic programming, and neural theorem proving) and motivates their roles in interactive compositionality.
Section~3 presents the AGEL-Comp architecture, detailing the Causal Program Graph world model, the planner--verifier loop, and the grounding mechanism.
Section~4 formalizes the perception--action--learning loop and the deduction--abduction cycle.
Section~5 introduces the Retro Quest evaluation protocol, experimental setup, and ablation studies, and analyzes the results.
Section~6 concludes with limitations and a forward-looking deployment outlook.

\section{Background}
Our framework integrates three key neuro-symbolic technologies. We briefly review each and motivate its inclusion.

\subsection{Causal Graphical Models}
A primary challenge for embodied agents is moving beyond statistical correlation to understand cause-and-effect. Causal graphical models, often represented as directed acyclic graphs (DAGs), are a formal tool for this purpose~\cite{cabalar2014causal,pearl2009causality,wu2025causal,fujitsu2024causal}. This structure moves beyond mere correlation to enable deeper reasoning, such as predicting the outcomes of interventions and contemplating counterfactuals \cite{s_s11,s_s8}. Our work draws inspiration from this paradigm by introducing the Causal Program Graph (CPG), which is specifically designed to represent an agent's procedural and operational knowledge.

\subsection{Inductive Logic Programming (ILP)}
ILP is a subfield of symbolic AI that induces a hypothesis $H$ (a logic program) that generalizes training examples $E$ given background knowledge $B$ ~\cite{cropper2022inductive,muggleton1994inductive}. In the standard setting of learning from entailment, a correct hypothesis $H$ must be both \textbf{complete} ($B \cup H \models E^+$) and \textbf{consistent} ($B \cup H \not\models E^-$) \cite{s_s31}. ILP is prized for its data efficiency and the interpretability of its symbolic outputs \cite{s_r10}. In neuro-symbolic systems, ILP provides a formal mechanism for rule induction from noisy, unstructured data, with applications in reinforcement learning and robotics \cite{s_s17,s_s31}. Unlike black-box neural networks that require vast amounts of data, ILP's strong inductive bias allows it to learn generalizable rules from very few examples. This sample efficiency is critical for an interactive agent that must learn robustly from sparse and often costly environmental feedback.

\subsection{Neural Theorem Provers (NTP)}
An NTP is an end-to-end differentiable relaxation of the backward-chaining algorithm used in classical theorem provers \cite{rocktaschel2017end,s_r6}. It operates on continuous vector representations of symbols (predicates and constants) and replaces discrete unification with a differentiable similarity function between embeddings \cite{s_r5}. The success of a proof is a differentiable score, allowing the entire system to be trained via gradient-based optimization. NTPs are valued for their ability to bridge symbolic reasoning with neural representation learning, providing interpretability and strong systematic generalization properties \cite{tsoukalas2024putnambench}. While computationally intensive, their scalability is an active area of research \cite{s_s19}.
An LLM planner is prone to generating plausible but logically unsound steps. We employ an NTP as a formal verifier to enforce compositional discipline. This is a more rigorous approach than simply prompting an LLM to ``check its own work," as the NTP provides a provable guarantee of a plan's validity with respect to the agent's learned world model.


\section{The AGEL-Comp Architecture}
\begin{figure}[ht!]
    \centering
    \includegraphics[width=1.0\linewidth]{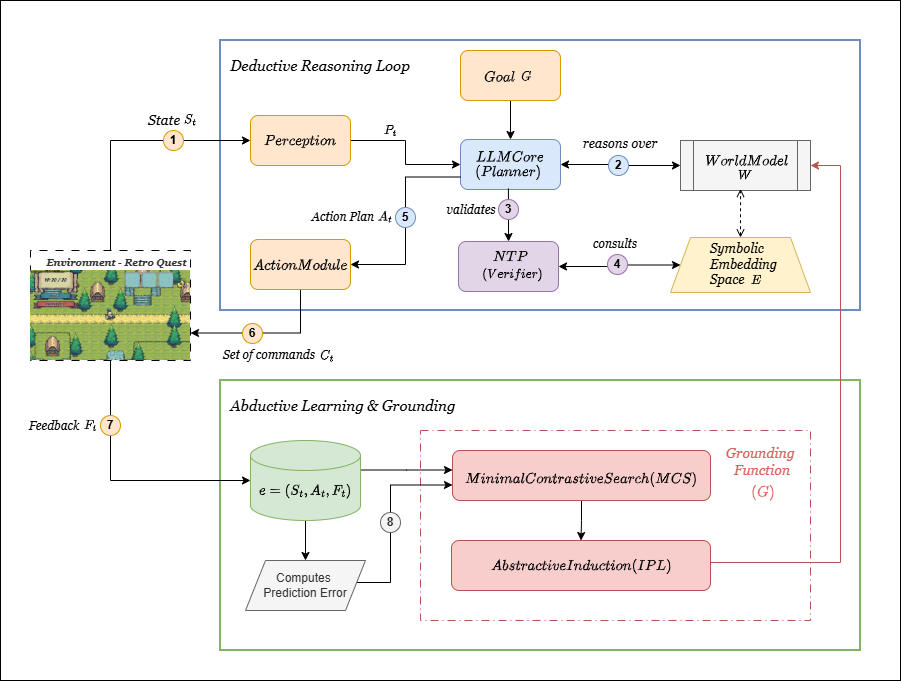}
    \caption{The AGEL-Comp neuro-symbolic architecture.}
    
    \label{fig:method}
\end{figure}

AGEL-Comp is a modular, neuro-symbolic architecture designed to learn a compositional world model from experience, as depicted in Figure \ref{fig:method}. It consists of the following interconnected components.

\subsection{Perception}
At each time step $t$, this module perceives the state of the simulated environment $S_t$ and translates this into a structured percept $P_t$. The percept $P_t$ is a set of ground literals representing the agent's current knowledge of the environment's entities and their states.

\subsection{LLM Core}
The LLM Core is the central cognitive engine, processing the percept $P_t$ in the context of a goal $G$. It consists of:

\subsubsection{World Model ($W$)}
The agent's stateful representation of the world is a causal directed hypergraph $W = (\mathcal{V}, \mathcal{E})$ representing knowledge as a structured, executable program. This structure makes procedural dependencies explicit.
\begin{itemize}
    \item \textbf{Nodes ($\mathcal{V}$)}: A set of grounded predicates (e.g., $is\_harmful(X)$) or concepts (e.g., `fire`).
    \item \textbf{Hyperedges ($\mathcal{E}$)}: A set of Horn clauses functioning as sub-programs. A hyperedge connects a set of input nodes (the clause body) to a single output node (the clause head), modeling a logical or causal dependency. For example, the rule $h \leftarrow b_1, \dots, b_n$ is a hyperedge from $\{b_1, \dots, b_n\} \subset \mathcal{V}$ to $h \in \mathcal{V}$.
\end{itemize}
This modular structure facilitates hierarchical planning and targeted model revision during learning.

\subsubsection{Goal Processing \& Planning}
This component is a hybrid planner-verifier architecture that integrates the LLM's generative capabilities with symbolic rigor.

\begin{enumerate}
    \item \textbf{LLM as Planner}: Given a goal $G$ and percept $P_t$, the LLM generates a set of candidate plans or sub-goals
    the LLM acts as a generative hypothesis engine that produces a set of candidate sub-goals,  $\{g_1, \dots, g_k\} = \text{LLM}_{generate}(G, P_t)$. This set represents a pool of plausible but unverified ideas. The NTP then processes this set, filtering for logical consistency and soundness.
    \item \textbf{NTP as Verifier}: Each sub-goal $g_i$ is a formal query for a NTP. It performs end-to-end differentiable backward chaining over the rules in $W_t$. It returns a tuple $(\sigma_i, \pi_i) = \text{NTP.prove}(g_i | W_t)$, where $\sigma_i \in $ is the proof success score and $\pi_i$ is the proof path. The path $\pi_i$ is then translated into an executable action sequence $A_t$.
\end{enumerate}
This loop ensures that plans are both creatively generated and logically sound according to the agent's grounded world model $W_t$.


\subsection{Action Module}
Translates the high-level textual action plan $A_t$ selected by the LLM core into a sequence of low-level commands $C_t$, directly executable via the simulation environment's API.

\subsection{Feedback Signals ($F$)}
For every executed action plan $A_t = (a_1, \dots, a_j)$, the agent receives a corresponding set of feedback signals $F_t = (f_1, \dots, f_j)$ from the environment. Each signal $f_i$ provides both the content (e.g., `HP: -10`) and intensity of an outcome, serving as the primary driver for learning.

\subsection{Episodic Memory ($M$)}
A fixed-size buffer that stores the $k$ most recent experiences. Each experience $e$ is a tuple: $e = (S_t, A_t, F_t)$, where $S_t$ is the state in which the action plan $A_t$ was executed, and $F_t$ is the resulting feedback. The full memory is $M = \{e_1, e_2, \dots, e_k\}$, which provides the data for the grounding process.

\subsection{Grounding Function ($\mathcal{G}$)}
A core challenge for any agent that learns from interaction is bridging the gap between low-level, noisy experiences and high-level, abstract symbolic knowledge. The Grounding Function ($\mathcal{G}$) in AGEL-Comp addresses this by decomposing the problem into two distinct, sequential stages: (1) a causal attribution stage to solve the credit assignment problem; and (2) an abstractive induction stage to generalize specific observations into reusable rules.

\subsubsection{Stage 1: Experience-Driven Causal Attribution}
The first stage addresses the credit assignment problem when a prediction error occurs---a mismatch between the outcome predicted by the agent's world model $W_t$ and the actual feedback $F_t$ received from the environment. To solve this credit assignment problem without confounds from the LLM's latent knowledge, we introduce a strict, experience-based algorithm: Minimal Contrastive Search (MCS). The MCS algorithm operates as follows:

\begin{enumerate}
    \item The agent executes an action plan $A_t$ in state $S_t$. The NTP, using the current world model $W_t$, predicts an outcome $F_{expected}$. The environment returns a different, unexpected feedback signal $F_{actual}$ (e.g., $F_{expected} = \text{HP: 0}$, 
    $F_{actual} = \text{HP: -10}$). This tuple $e_{fail} = (S_t, A_t, F_{actual})$ 
    is the trigger for learning.

    \item The perception module provides a set of ground literals $C = \{c_1, \dots, c_n\}$ representing the trigger state $S_t$. For example: $C$ = \{\texttt{at(agent, loc1)}, \texttt{is(fire, loc1)}, \texttt{is(coin, loc1)}\}.

    \item The MCS module searches the agent's episodic memory $M$ for a "minimal pair" experience, $e_{success}$. A valid $e_{success}$ must meet two criteria: 
    (a) it involved the \textit{same action} $A_t$ (or a semantically equivalent 
    one) and (b) it resulted in the \textit{expected} outcome $F_{expected}$.

    \item If a valid $e_{success}$ is found, the module compares the state literals $C_{fail}$ (from $S_t$) and 
    $C_{success}$ (from the state of $e_{success}$). It computes the state 
    difference: $\Delta S = C_{fail} - C_{success}$.

    \item The algorithm proceeds only if 
    $\Delta S$ contains \textit{exactly one} literal (a "singleton"). If 
    $\Delta S = \{c_k\}$, this single literal (e.g., $\texttt{is(fire, loc1)}$) 
    is identified as the most probable causal antecedent for the failure. The 
    module then formulates a specific, grounded causal hypothesis, $h_{causal}$ 
    (e.g., $\texttt{causes\_damage(fire)}$).
\end{enumerate}

\subsubsection{Stage 2: Abstractive Induction via Meta-Interpretive Learning}
This stage takes the specific hypothesis $h_{causal}$ (if one was found) and 
generalizes it into a reusable symbolic rule using a Meta-Interpretive 
Learning (MIL) system.

\begin{enumerate}
    \item If the MCS algorithm returned 
    $\texttt{null}$ (see Section 3.6.3), the learning cycle is skipped. If it 
    returned a valid $h_{causal}$, this literal becomes the positive example, 
    $E^+$, for the MIL engine. The existing CPG, $W_t$, serves as background 
    knowledge, $B$.

    \item  The MIL engine uses higher-order 
    metarules to search for a general hypothesis $H$ (a Horn clause) that 
    explains $E^+$ given $B$. For example, given 
    $E^+ = \text{causes\_damage(fire)}$ and background knowledge 
    $B = \{\text{is\_harmful(fire)}\}$, the engine can induce the rule 
    $H = \forall X: \text{causes\_damage}(X) \leftarrow \text{is\_harmful}(X)$.

    \item This new, empirically-grounded rule 
    $H$ is integrated into the CPG: $W_{t+1} \leftarrow W_t \cup H$. 
    This rule is now available to the NTP for all future planning cycles.
\end{enumerate}





\subsection{Integration of NTP \& CPG}
The synergy between the symbolic CPG and the neural NTP is enabled by two key processes: a shared, trainable embedding space for all symbols, and a dual-phase training regimen for the NTP.

\subsubsection{Symbol Embedding Space}
The bridge between the discrete, symbolic representation of the CPG and the continuous vector space of the NTP is a trainable embedding matrix $\mathbf{E} \in \mathbb{R}^{|\mathcal{V}| \times d}$, where $|\mathcal{V}|$ is the size of the predicate and constant vocabulary in the CPG and $d$ is the embedding dimension. Each symbolic predicate or constant $v \in \mathcal{V}$ is mapped to a unique dense vector $\mathbf{v} \in \mathbb{R}^d$.

This shared embedding space allows the NTP to perform its core operation: differentiable unification. Instead of a discrete check for identity, the NTP computes a soft unification score between a goal's embedding and a rule head's embedding based on the similarity (e.g., dot product) of their vector representations \cite{rocktaschel2017end,s_r6}. When new predicates are invented by the ILP engine, they are added to the vocabulary $\mathcal{V}$, and their corresponding embeddings are initialized in $\mathbf{E}$ (e.g., randomly or by composing the embeddings of their constituent parts) before being refined during training.

\subsubsection{NTP Training Regimen}
The NTP and the symbol embedding matrix $\mathbf{E}$ are not static; they are trained to adapt to the agent's growing knowledge base. The training follows a two-phase process:

\begin{enumerate}
    \item \textbf{Phase 1: Bootstrapping on Foundational Knowledge.} Initially, the NTP and $\mathbf{E}$ are pre-trained on a set of foundational, domain-general logical rules (e.g., transitivity, symmetry) and basic facts about the environment. The training objective is a link-prediction task: to maximize the proof success score $\sigma$ for known true facts while minimizing it for false or unknown facts. This phase ensures that the symbol embeddings are initialized to a meaningful state and that the NTP learns the fundamental mechanics of logical deduction before encountering more complex, learned rules.
    
    \item \textbf{Phase 2: Continual Fine-tuning on Induced Rules.} As the agent interacts with the world, the ILP engine induces new rules $\Delta W$, which are added to the CPG to form $W_{t+1}$. The NTP's knowledge must be updated to reflect this new understanding. Therefore, the agent periodically performs fine-tuning steps on the NTP and the embedding matrix $\mathbf{E}$ using the updated CPG, $W_{t+1}$. This is done by sampling queries that can be proven with the new rules and optimizing the same link-prediction objective. This continual learning process ensures that the NTP's reasoning remains consistent with the agent's latest empirically grounded knowledge, solidifying the integration between the symbolic and neural components.
\end{enumerate}
This dual-phase training regimen allows the NTP to serve as a robust and adaptable verifier that evolves in lockstep with the agent's symbolic world model.

\section{AGEL-Comp Workflow and Learning Mechanism}
The agent's operational loop, detailed in Algorithm \ref{algo_modular}, alternates between two phases: \emph{acting} and \emph{learning}.

The acting phase (\texttt{Perceive\_And\_Act}) begins with the agent observing its environment, $S_t$. For example, a \texttt{shiny\_coin} is next to a \texttt{crackling\_fire}. To achieve its goal $G$ (\texttt{"Retrieve the shiny coin!"}), the LLM core consults the current World Model, $W_t$. If $W_t$ contains no rules about fire being dangerous, the LLM may propose a direct plan, $A_t$ (\texttt{$a_1$: approach(fire); $a_2$: grab(coin)}). The NTP verifies this plan as valid against $W_t$. The agent executes the plan and receives negative feedback ($f_1$: a decrease in HP, \texttt{damage=-50HP}). The \textit{interactions} in this attempt is 1 (only got to first action). This experience is captured as a set of episodes, which are then returned to the main operational loop to be recorded in episodic memory, $M$.

This observed outcome constitutes a prediction error, as the negative feedback was not anticipated by the current world model $W_t$. This error triggers the learning phase (\texttt{Learn\_From\_Experience}). The grounding function ($G$), using the principled mechanism of Causal Attribution and Abstractive Induction, analyzes the episode to understand the cause of the error and repair the agent's world model. It induces a set of Horn clauses ($\Delta W$) that explain the event:
\begin{itemize}
    \item \texttt{causes\_damage(X) :- is\_harmful(X).}
    \item \texttt{is\_harmful(fire).}
\end{itemize}

This rule set is then integrated into the World Model ($W_{t+1} \leftarrow W_t \cup \Delta W$). This cycle enables the agent to continuously refine its understanding by grounding symbolic knowledge in direct experience.

\begin{algorithm}[ht!]
\caption{AGEL-Comp: Perception-Action-Learning Loop}
\label{algo_modular}
\begin{algorithmic}[1]

\Function{Perceive\_And\_Act}{$G, W$}
\Statex \Comment{\emph{Helper function to decide on and execute an action}}
    \State $S_t \gets \textbf{Perception.Observe()}$
    \State $A_t \gets \textbf{LLM\_Core.Plan}(G, S_t, W)$
    \State $C_t \gets \textbf{ActionModule.Translate}(A_t)$
    \State $F_t \gets \textbf{Environment.Execute}(C_t)$
    \State \textbf{return} $\{(S_t, a_1, f_1)...(S_t, a_j, f_j)\}$ \Comment{Return episodes}
\EndFunction
\Statex 

\Function{Learn\_From\_Experience}{$M, W$} 
\Statex \Comment{\emph{Helper function to learn from past experiences}}
    \State $e_k \gets M.\textbf{GetRecentEpisodes}()$ 
    \State $\Delta W \gets \mathcal{G}(e_k)$ \Comment{Induce new rule(s) via Grounding function}
    \If{$\Delta W$ is not null}
        \State $W.\textbf{AddRules}(\Delta W)$ \Comment{Update World Model ($W \cup \Delta W$)}
    \EndIf
\EndFunction
\Statex \rule{\linewidth}{0.5pt}

\State \textbf{Initialize} World Model $W$, Episodic Memory $M$, goal $G$
\Statex \Comment{\emph{Main Agent Loop}}
\While{agent is active}
    \State \Comment{Agent interacts with the world}
    \State $episodes \gets \textproc{Perceive\_And\_Act}(G, W)$
    \State $M.\textbf{Record}(episodes)$ \Comment{Adds new episodes to $M$}
    
    \State \Comment{Agent reflects and updates its understanding}
    \If{it is time to learn}
        \State \textproc{Learn\_From\_Experience}(M, W)
    \EndIf
\EndWhile
\end{algorithmic}
\end{algorithm}

This closes the loop. The next time the agent plans, the NTP will consult the updated World Model $W_{t+1}$. A plan to approach the fire would now be evaluated by the NTP. When evaluating a plan involving \texttt{approach(fire)}, the core would query the NTP to prove the expected outcome. The NTP would find a proof for the goal $\texttt{causes\_damage(agent)}$ via the path: $\texttt{is\_harmful(fire)}$ is a known fact, and the rule $\texttt{causes\_damage(X) :- is\_harmful(X)}$ connects it to the negative outcome. Since this proof leads to a known negative outcome, the NTP would return a very low proof success score ($\sigma \approx 0$), causing the plan to be rejected and forcing the LLM to generate a safer alternative. 

AGEL-Comp's learning is a hybrid process with two complementary forms. First, new knowledge is captured through the symbolic addition of new Horn clauses ($\Delta W$) to the world model. Subsequently, neural reasoner adaptation occurs as the NTP and the symbol embedding matrix $E$ are periodically fine-tuned on this updated knowledge base. This synergy keeps the agent's deductive capabilities tightly aligned with its evolving, empirically-grounded symbolic understanding of the world.

\section{Experiments}
The core idea of this work is that the AGEL-Comp framework enables agents to overcome the inherent compositional limitations of LLMs by facilitating in-situ, grounded learning. To test this, we designed an evaluation protocol that measures zero-shot generalization within a single, continuous run, without any offline model training or fine-tuning. Our experiments are designed not only to validate this claim against a baseline but also to dissect the contribution of each component within our framework through rigorous ablation studies.

\subsection{Experimental Hypotheses}
We structure our evaluation to test three primary hypotheses:
\begin{itemize}
    \item[\textbf{H1:}] The AGEL-Comp framework will outperform a standard LLM-based agent in environments with ambiguous or stochastic events.
    \item[\textbf{H2:}] Ablated versions of AGEL-Comp lacking either the NTP (verifier) or the ILP engine (learner) will show a decrease in performance and efficiency:
    \begin{enumerate}
        \item[2a] The agent without the NTP verifier (w/o NTP) will have a degraded performance as compared to the full system with the NTP verifier.
        \item[2b] The agent without the ILP learner (w/o ILP) will fail at generalizing to novel tasks that require new knowledge, resulting in a lower performance as compared to the full system with the ILP learner.
    \end{enumerate}
\end{itemize}

\subsection{Simulation Environment-- Retro Quest}
To evaluate our agent in a rich, interactive world, we developed \texttt{Retro Quest for Compositionality}, a top-down 2D action RPG game. The environment was built using the Unity engine and the ML-Agents toolkit, which provides the interface for agent control and observation. The visual assets are based on the ``Tiny Swords'' pixel art package by Pixel Frog\footnote{\url{https://pixelfrog-assets.itch.io/tiny-swords}}, giving the world a classic RPG aesthetic with characters, monsters (e.g., goblins), NPCs (e.g., farmers), and interactive objects (e.g., sheep, pumpkins). The agent perceives the world from a top-down perspective, receiving both visual information and a structured list of nearby entities and their states. Its action space includes navigation, basic attacks, and interaction with objects and NPCs (e.g., move\_right, talk, pickup, use\_item). 

\subsection{Models and Configurations}
We compare four distinct agent configurations using the same underlying multimodal LLM as the core cognitive engine, as described in Table~\ref{tab:agent_configs}.
\begin{table}[ht!]
    \centering
    \caption{Agent Configurations for Experimental Evaluation and Ablation Studies.}
    \label{tab:agent_configs}
    \begin{tabularx}{\columnwidth}{lXc}
        \toprule
        \textbf{Configuration} & \textbf{Description} \\
        \midrule
        \textbf{MLLM-Agent (Baseline)} & A standard ReAct-style agent with no persistent, explicit world model. \\
        \addlinespace 
        \textbf{AGEL-Comp (Full System)} & Our complete proposed architecture with the CPG, confidence-aware ILP learner, and NTP verifier. \\
        \addlinespace
        \textbf{AGEL-Comp (w/o NTP)} & An ablated version that learns new rules via ILP but has no NTP verifier. Plans are executed based on the LLM's judgment alone. This tests the value of logical verification. \\
        \addlinespace
        \textbf{AGEL-Comp (w/o ILP)} & An ablated version with an NTP verifier but a static, pre-loaded world model. It cannot learn new rules from experience. This tests the value of the grounding and induction mechanism. \\
        \bottomrule
    \end{tabularx}
\end{table}

We use 4 state-of-the-art multimodal models as the backbone for both configurations: GPT-4o\footnote{\url{https://openai.com/index/hello-gpt-4o/}}, Gemini Pro 2.5\footnote{\url{https://ai.google.dev/gemini-api/docs/models#gemini-2.5-pro}}, LLaVA 1.6\footnote{\url{https://huggingface.co/llava-hf/llava-v1.6-mistral-7b-hf}}, and Deepseek v1\footnote{\url{https://huggingface.co/deepseek-ai/deepseek-vl-7b-chat}}.

Each agent configuration is executed through a 10-quest curriculum in a single, continuous session, with random seeds used three times. The quests are designed to test compositional generalization, with later quests explicitly incorporating ambiguous scenarios to probe for robust, grounded learning. Before conducting the experiments, we pre-registered our protocol on the Open Science Framework (OSF) at: https://osf.io/a6j4c.




\subsection{Metrics}
We use a set of metrics, as noted in Table~\ref{tab:metrics}, designed to capture an agent's ability to learn and generalize on the fly. {Note that, while most of the metrics apply to all agent configurations, some metrics like Adaptations Trials and Rules Learned are specific to AGEL-Comp.

\begin{table}[ht!]
    \centering
    \caption{Evaluation Metrics}
    \label{tab:metrics}
    \begin{tabularx}{\columnwidth}{lX}
        \toprule
        \textbf{Metric} &  \textbf{Description} \\
        \midrule
        \textbf{Quest Success Rate (\%)} & Measure the proportion of successfully completed quests. Failures include incomplete quests or crashes.\\
        \textbf{First-Try Success Rate (\%)}  & Measures zero-shot generalization and the framework's ability to produce a correct plan without trial-and-error. \\
        \textbf{Iterations}  & Measures the total number of iterations the framework takes in order to succeed in the quest. \\
        \textbf{Sample Efficiency}  & Measures the total number of interactions the system made to reach a performance threshold. \\
        \textbf{Adaptation Trials} & Measures how quickly an agent learns the correct rule after a failure.  \\
        \textbf{Rules Learned}  & Measures the number of rules generated by ILP and integrated into the CPG. \\
        \bottomrule
    \end{tabularx}
\end{table}


\subsection{Results and Analysis}
\label{sec:results}

Our experimental results, summarized in the aggregated data (Table~\ref{tab:results}) and visualized in Figures~\ref{fig:success}, \ref{fig:success_LLMs}, \ref{fig:iterations_completetion}, and \ref{fig:slope_chart}, provide strong support for our primary hypotheses. The data clearly indicates that the AGEL-Comp architecture (H1) significantly outperforms the MLLM-Agent baseline and that its core components (H2)---the ILP learner and the NTP verifier---are both critical and synergistic.

\begin{sidewaystable} 
\centering
\caption{Aggregated Experiment Results (Mean ± Std)}
\label{tab:results}
\begin{tabular}{llrrrrr} 
\toprule
         LLM &         Agent Config & Quest Success (\%) & First Try (\%) &     Avg. Iters. &       Avg. Time (s) & Avg. Sample Eff. \\
\midrule
        GPT-4o &    MLLM-Agent-Baseline &         86.67 ± 5.77 &     6.67 ± 5.77 &    18.87 ± 1.09 &     347.88 ± 41.69 &    146.97 ± 12.88 \\
        GPT-4o & AGEL-Comp-Full-System &        100.00 ± 0.00 &    66.67 ± 5.77 &    11.38 ± 0.56 &       36.40 ± 3.06 &     25.03 ± 3.43 \\
        GPT-4o &       AGEL-Comp-w/o-NTP &        100.00 ± 0.00 &    23.33 ± 5.77 &    13.72 ± 0.43 &     104.67 ± 13.30 &     62.10 ± 10.31 \\
        GPT-4o &       AGEL-Comp-w/o-ILP &         86.67 ± 5.77 &    10.00 ± 0.00 &    18.89 ± 0.35 &     340.40 ± 61.38 &    159.57 ± 32.28 \\
\addlinespace
Gemini-2.5-Pro &    MLLM-Agent-Baseline &         83.33 ± 5.77 &     3.33 ± 5.77 &    17.73 ± 1.27 &     302.02 ± 30.40 &    159.30 ± 21.78 \\
Gemini-2.5-Pro & AGEL-Comp-Full-System &        100.00 ± 0.00 &    66.67 ± 5.77 &    10.81 ± 2.44 &       28.06 ± 7.11 &     23.27 ± 5.00 \\
Gemini-2.5-Pro &       AGEL-Comp-w/o-NTP &         93.33 ± 5.77 &    20.00 ± 0.00 &    12.82 ± 0.74 &     108.36 ± 22.16 &     72.63 ± 16.13 \\
Gemini-2.5-Pro &       AGEL-Comp-w/o-ILP &         83.33 ± 5.77 &    10.00 ± 0.00 &    16.96 ± 1.77 &     269.13 ± 58.66 &    142.67 ± 24.96 \\
\addlinespace
DeepSeek-VL-7B &    MLLM-Agent-Baseline &         66.67 ± 5.77 &     3.33 ± 5.77 &    20.36 ± 0.66 &     739.81 ± 37.12 &    243.00 ± 30.24 \\
DeepSeek-VL-7B & AGEL-Comp-Full-System &        100.00 ± 0.00 &    56.67 ± 5.77 &    15.10 ± 0.61 &       74.36 ± 19.01 &     40.43 ± 6.39 \\
DeepSeek-VL-7B &       AGEL-Comp-w/o-NTP &         86.67 ± 5.77 &    23.33 ± 5.77 &    16.49 ± 0.64 &     246.78 ± 21.10 &    115.83 ± 14.47 \\
DeepSeek-VL-7B &       AGEL-Comp-w/o-ILP &         70.00 ± 10.00 &     6.67 ± 5.77 &    19.43 ± 0.75 &     595.85 ± 60.95 &    223.97 ± 19.91 \\
\addlinespace
       LLaVA-1.6 &    MLLM-Agent-Baseline &         63.33 ± 5.77 &     0.00 ± 0.00 &    22.95 ± 0.43 &    1028.72 ± 152.55 &    258.10 ± 43.30 \\
       LLaVA-1.6 & AGEL-Comp-Full-System &        100.00 ± 0.00 &    50.00 ± 10.00 &    17.22 ± 0.92 &       96.81 ± 20.32 &     41.70 ± 7.39 \\
       LLaVA-1.6 &       AGEL-Comp-w/o-NTP &         86.67 ± 5.77 &    23.33 ± 5.77 &    18.43 ± 1.14 &     378.13 ± 14.71 &    126.17 ± 6.50 \\
       LLaVA-1.6 &       AGEL-Comp-w/o-ILP &         66.67 ± 11.55 &     6.67 ± 5.77 &    21.33 ± 0.75 &     876.12 ± 74.87 &    246.83 ± 22.29 \\
\bottomrule
\end{tabular}
\end{sidewaystable} 

\subsubsection{H1: AGEL-Comp vs. Baseline}

The AGEL-Comp system demonstrates superior performance, robustness, and efficiency compared to the standard MLLM-Agent baseline.

\begin{itemize}
    \item \textbf{Overall Performance:} The AGEL-Comp (Full) system achieved a \textbf{100\%} Quest Success Rate across all four LLM backbones (Figure~\ref{fig:success}). The baseline agent's performance was not only lower but also highly dependent on the LLM, ranging from 86.67\% (GPT-4o) down to 63.33\% (LLaVA-1.6) (Figure~\ref{fig:success_LLMs}. As shown in the performance degradation (Figure.~\ref{fig:slope_chart}), the baseline and \texttt{w/o ILP} models suffer a catastrophic drop in success on ``Hard" (Level 4) and ``Very Hard" (Level 5) quests, falling to 0\% success. In contrast, the AGEL-Comp system maintains 100\% success even on the most difficult quests, proving it can overcome the compositional challenges that crashed the baseline.

    \item \textbf{Zero-Shot Generalization (First-Try Success):} As shown in Figure~\ref{fig:success} (right). The AGEL-Comp system achieved a mean success of \textbf{60.0\%}, while the baseline agent was \textit{18 times worse} at \textbf{3.3\%}. As per Figure~\ref{fig:slope_chart}, we can observe the baseline's reliance on statistical pattern matching, which fails when faced with novel (but compositionally simple) problems. AGEL-Comp's high first-try success (in Figure ~\ref{fig:success_LLMs}), shows that its hybrid reasoning core—where the NTP verifies LLM-proposed plans against the CPG and ILP learning new rules and updating the world model—allows it to \textit{deduce} a correct, novel plan \textit{before} acting, demonstrating true compositional reasoning.

    \item \textbf{Efficiency:} The heatmaps in Figure.~\ref{fig:iterations_completetion} show the baseline and \texttt{w/o ILP} agents taking progressively more iterations and time on later quests. The AGEL-Comp system remains highly efficient. This is confirmed by the \textbf{Average Sample Efficiency} (Table~\ref{tab:results}). For example, with Gemini-2.5-Pro, the AGEL system required only 23.27 samples on average, whereas the baseline required 159.30---a \textbf{6.8$\times$ improvement} in sample efficiency, as it learns from a single failure rather than repeating it.
\end{itemize}

\begin{figure}[ht!]
    \centering
    \includegraphics[width=1.0\linewidth]{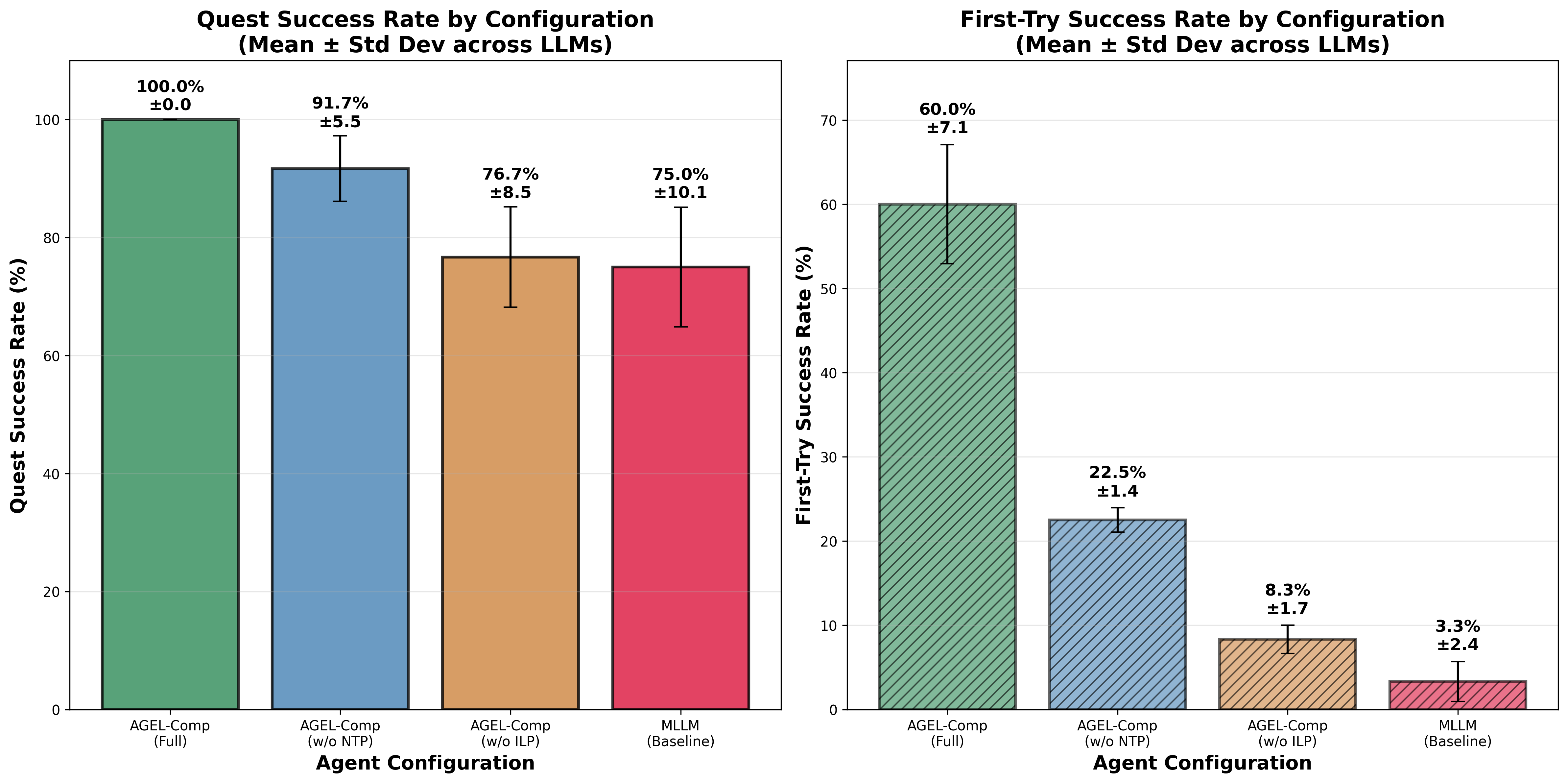}
    \caption{Aggregated Quest Success and First-Try Success Rate by Agent Config.}
    \label{fig:success}
\end{figure}

\begin{figure}[ht!]
    \centering
    \includegraphics[width=1.0\linewidth]{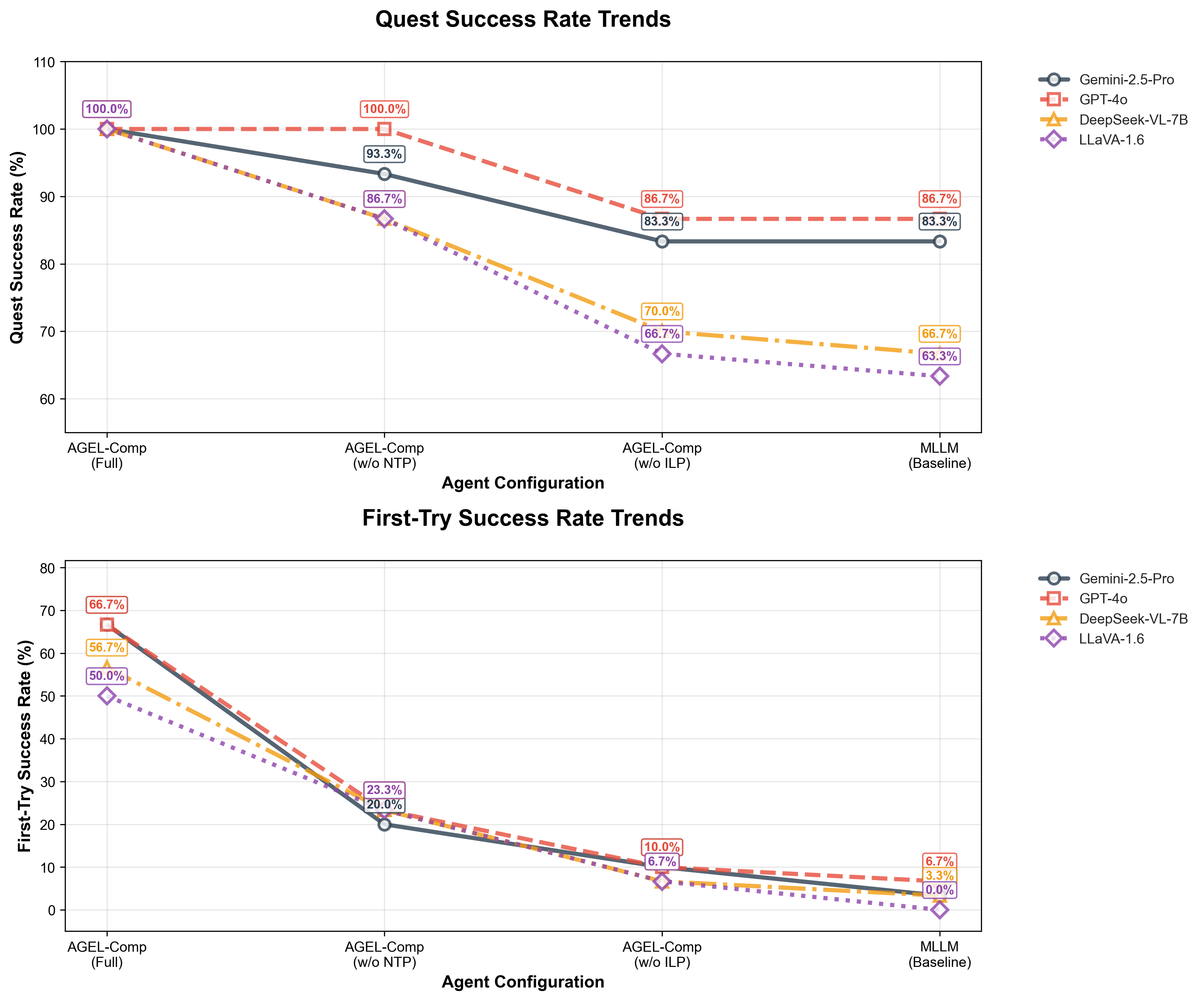}
    \caption{Aggregated Quest Success and First-Try Success Rate Per LLM Per Agent Config.}
    \label{fig:success_LLMs}
\end{figure}

\begin{figure}[ht!]
    \centering
    \includegraphics[width=1.0\linewidth]{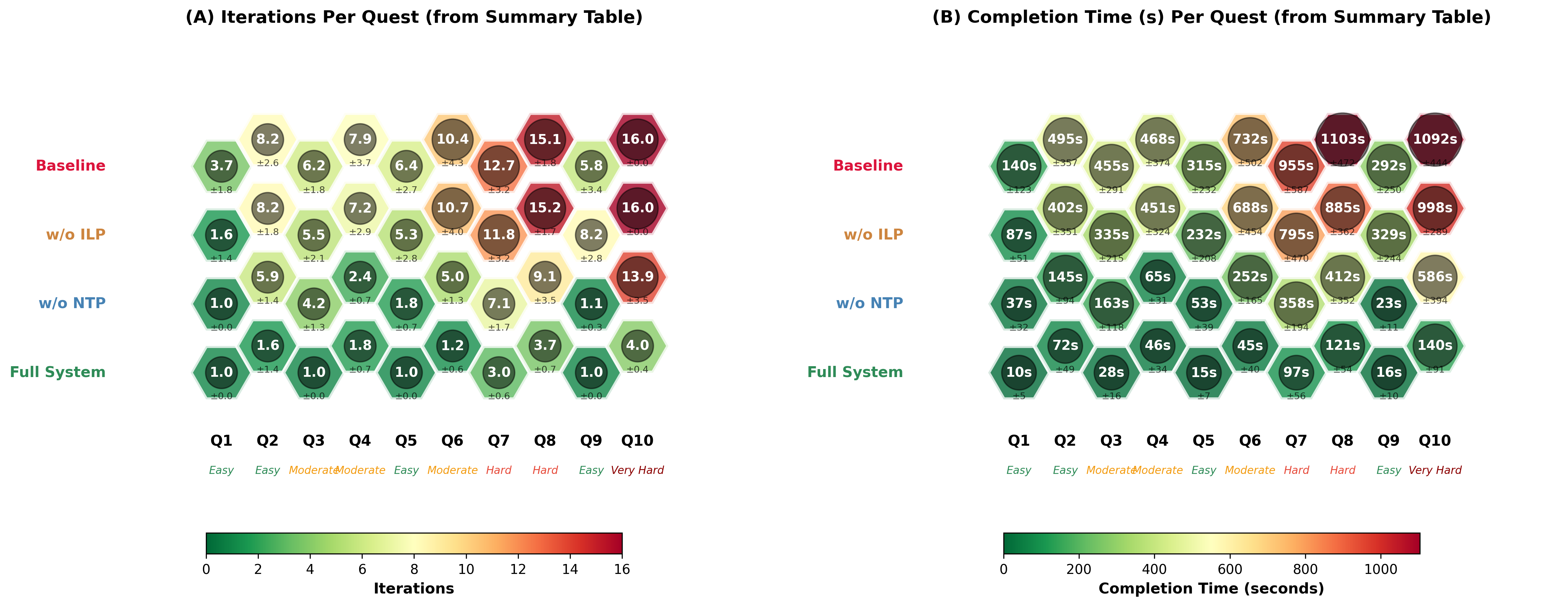}
    \caption{Aggregated Iterations and Completion Time Per Agent Config. Per Quest}
    \label{fig:iterations_completetion}
\end{figure}
\paragraph{\textbf{Discussion}} The baseline agent's failure, particularly in First-Try Success, confirms our premise: pure LLM-based agents lack the grounded, systematic understanding needed for robust interaction. They are brittle and rely on trial and error. AGEL-Comp's success even on the hardest quests, shows it successfully learns and applies the compositional "grammar" of the environment. 

Interestingly, while the full AGEL-Comp system achieved 100\% success with all models, the baseline's performance revealed a gap between model classes (Figure ~\ref{fig:success_LLMs}). The closed-source models (GPT-4o, Gemini-2.5-Pro) showed more robust baseline performance than the open-source models (DeepSeek-VL, LLaVA-1.6). This highlights a practical trade-off: stronger, closed-source models provide a better foundation, but this advantage is often tied to API dependency, token costs, and latency, reinforcing the value of our framework that can maximize the capabilities of more accessible open-source alternatives.

\subsubsection{H2: Ablation Study and Component Synergy}

The ablation studies confirm H2a and H2b, revealing that the ILP learner and NTP verifier are not just additive but create a necessary synergy.

\begin{itemize}
    \item \textbf{H2b: The \texttt{w/o ILP} Agent (Verifier, No Learner):} This ablation confirms the necessity of the inductive learning mechanism. This agent, which has a verifier but a static world model, performed almost identically to the baseline (Figures ~\ref{fig:success} \& ~\ref{fig:success_LLMs}). Its Quest Success (76.7\% mean) and First-Try Success (8.3\% mean) were abysmal. As seen in Figure~\ref{fig:slope_chart}, it fails completely on hard tasks. This is a critical insight: a logical verifier (NTP) is useless if the underlying world model ($W$) is incomplete or wrong. The agent simply repeats its "logically valid" but factually incorrect plan, failing to learn from experience.

    \item \textbf{H2a: The \texttt{w/o NTP} Agent (Learner, No Verifier):} This ablation confirms the value of logical verification. This agent, which can learn via ILP but executes any plan the LLM proposes without verification, performed 9\% lower than the AGEL system in quest success and drastically lower by dropping from 60\% to 22.5\% on first-try success (Figures~\ref{fig:success} \& ~\ref{fig:success_LLMs}). This agent learns ``the hard way"---by executing a flawed plan, receiving negative feedback, and \textit{then} inducing a new rule. The  AGEL-Comp system \textit{avoids} this failure by using the NTP to reject the flawed plan \textit{before} execution, forcing the LLM to find a safer alternative on the first try.
\end{itemize}

\paragraph{\textbf{Discussion}} The ablation results provide the paper's most critical insight. They demonstrate that logical reasoning and symbolic learning are symbiotic and mutually essential. The \texttt{w/o ILP} agent can only deduce, but its deductions are based on a faulty model. The \texttt{w/o NTP} agent can abductively learn, but it cannot safely deduce, leading to inefficient and risky trial-and-error. The AGEL-Comp system is the only one that closes the loop: it deduces sound plans (via NTP) and abductively repairs its model when those deductions fail (via ILP), enabling it to learn the ``grammar" of its world, as visualized by the CPG Growth in Figure ~\ref{fig:cpg_growth}. The non-zero rules retracted line further shows the system's robustness, as it self-corrects invalid knowledge.

\begin{figure}[ht!]
    \centering
    \includegraphics[width=1.0\linewidth]{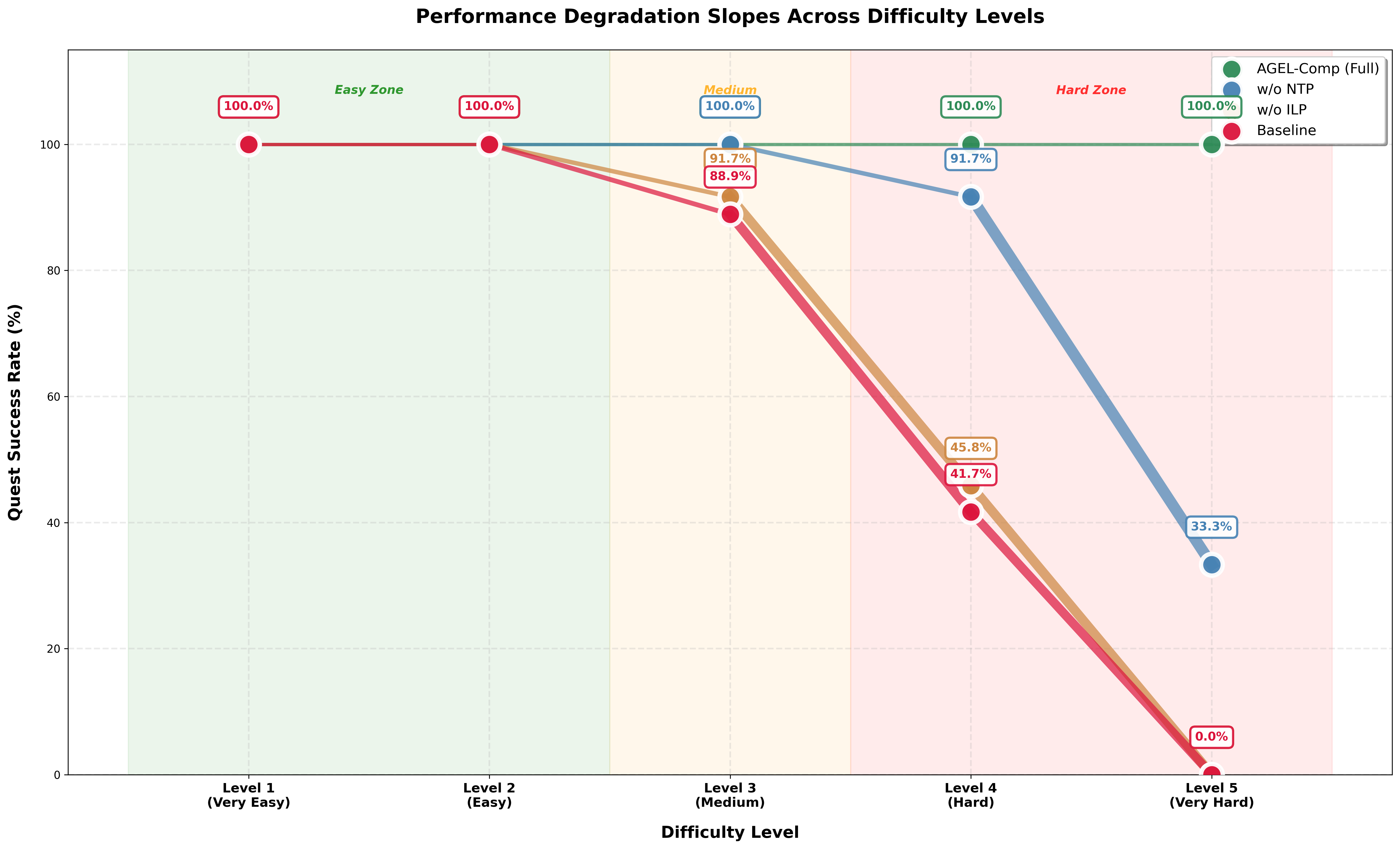}
    \caption{Performance degradation from easy to hard quests}
    \label{fig:slope_chart}
\end{figure}

\begin{figure}[ht!]
    \centering
    \includegraphics[width=1.0\linewidth]{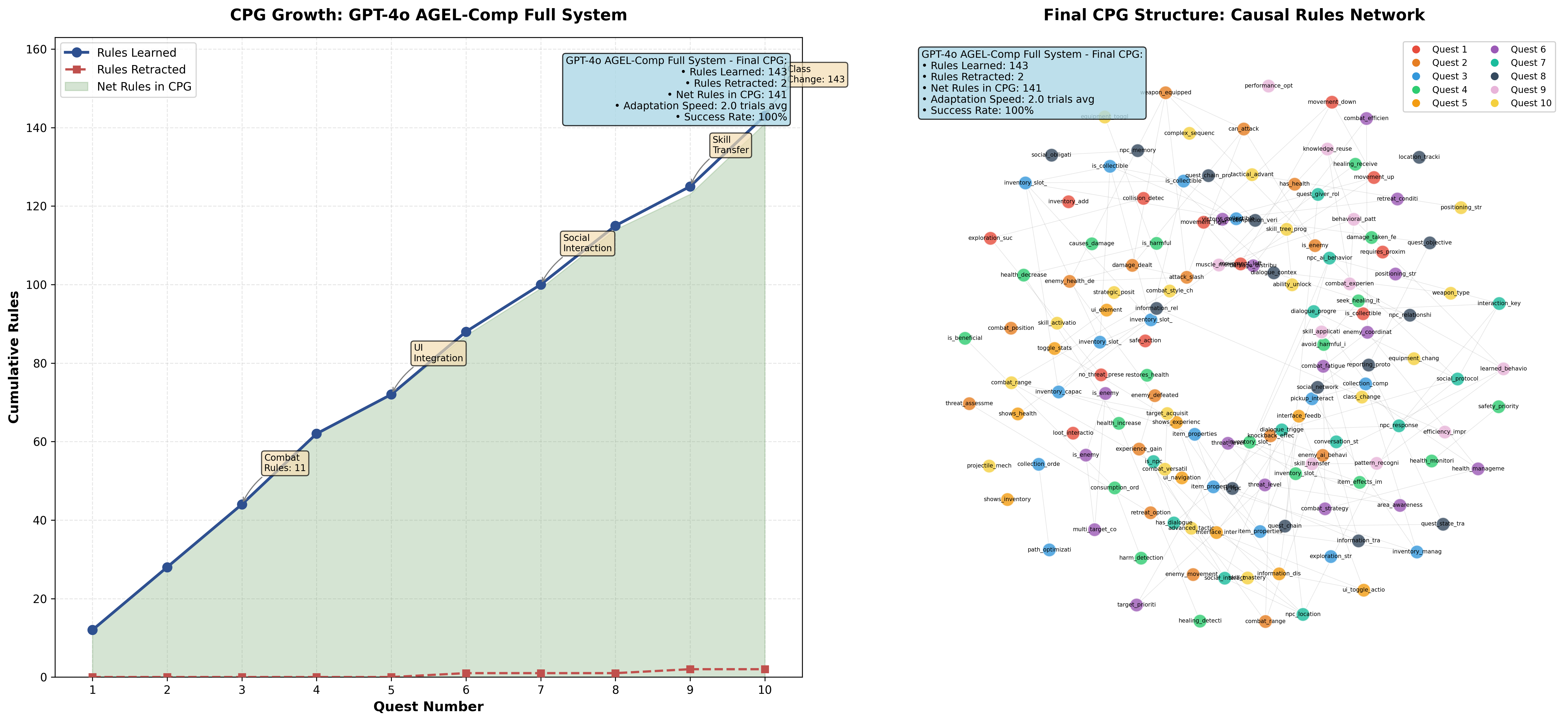}
    \caption{Development and Growth of Causal Program Graph by AGEL (GPT-4o)}
    \label{fig:cpg_growth}
\end{figure}

\subsection{Comparison to State-of-the-Art Neuro-Symbolic Systems}


AGEL-Comp sits at the intersection of three prominent neuro-symbolic lines: (i) differentiable logic and neural reasoning, (ii) symbolic program induction, and (iii) LLM-based agent architectures.
First, neuro-symbolic systems typically combine neural perception with (largely) fixed logical structure, using differentiable reasoning or constraint losses to enforce consistency, but they rarely \emph{expand} an executable rule base online (e.g., \cite{manhaeve2018deepproblog,rocktaschel2017end,serafini2016logic,yi2018nsvqa}), AGEL-Comp maintains a dynamic program-structured world model (CPG) as \emph{mutable} and learns new executable rules online from interaction. 
Second, compared to program induction and neuro-symbolic learning approaches
AGEL-Comp is explicitly \emph{experience-driven}: symbolic rule acquisition is triggered by prediction errors during embodied interaction. The grounding function decomposes this into (a) causal attribution via Minimal Contrastive Search (to isolate a likely causal antecedent under minimal confounding) and (b) abstractive induction via meta-interpretive learning to generalize into reusable clauses. This yields sample-efficient symbolic learning under sparse feedback, rather than relying on large supervised corpora.
Third, relative to LLM-agent stacks that rely on prompting, tool use, and unstructured memory/self-reflection (e.g., \cite{ahn2022saycan,shinn2023reflexion,yao2022react}), AGEL-Comp enforces a planner--verifier separation in which LLM-proposed subgoals are checked against an explicit symbolic model via neural theorem proving, reducing plausible-but-invalid plans under distribution shift.
These three key architectural innovations directly targets compositional generalization in interactive settings by ensuring that new combinations of known primitives are mediated by explicit, reusable structure rather than latent statistical associations alone.

\subsubsection{Scalability Considerations}
The integration of multiple symbolic components in AGEL-Comp raises important questions of scalability. 
NTP verification can be computationally intensive; however, their use in our framework is constrained. It is not solving open-ended queries but verifying specific LLM-proposed sub-goals against the world model $W_t$. 
ILP induction, i.e., the search space for the meta-interpretive learning (MIL) engine, though potentially vast, is heavily constrained by the learning context. The abstractive induction stage is triggered only after the causal attribution module has isolated a single, specific causal literal (e.g., \texttt{causes\_damage(fire)}) to serve as a positive example, $E^+$, along with a small set of negative examples, $E^-$. This highly focused learning task—generalizing from one or two examples—dramatically prunes the hypothesis search space, making induction efficient.

This is clearly evident in Figure~\ref{fig:iterations_completetion}, directly addressing scalability concerns: while a single, verified AGEL-Comp iteration is computationally more intensive, this overhead is trivial compared to the drastic reduction in total iterations; the baseline agent makes many mistakes and has to retry, which adds up more iterations and makes it 5 to 10 times slower to finish and computationally costing more.

\paragraph{Outlook and practical deployment challenges.}
Real-world deployment will require (i) budgeting verification/induction cost (depth limits, caching, safety-only checks), (ii) robust symbol grounding under noisy/partial perception, (iii) controlling CPG growth and inconsistency (pruning and rule retraction), and (iv) safe adaptation under distribution shift via detection and repair of invalidated rules. We view these as key directions toward sim-to-real embodied agents.
\section{Conclusion}
\label{sec:conclusion}

Large Language Models (LLMs) fundamentally struggle with compositional generalization. This work introduced AGEL-Comp, a neuro-symbolic framework designed to solve this challenge by integrating the generative flexibility of LLMs with the formal rigor of symbolic reasoning. Our architecture is built on three pillars: a causal program graph (CPG) as a dynamic world model, an inductive logic programming (ILP) engine to ground new rules from experience, and a neural theorem prover (NTP) to verify plans for logical consistency.

Our experiments in the \texttt{Retro Quest} environment demonstrate that this hybrid approach is highly effective. The AGEL-Comp system achieved perfect quest success across all LLM backbones, even on very hard quests designed to cause compositional failure. In contrast, a standard LLM-agent baseline and ablated versions of our framework experienced a catastrophic collapse in performance when faced with the same novel scenarios.

The most crucial insight comes from our ablation studies. We show that neither symbolic learning (ILP) nor symbolic reasoning (NTP) is sufficient on its own. An agent with only a verifier (\texttt{w/o ILP}) fails because it cannot correct its flawed world model. An agent with only a learner (\texttt{w/o NTP}) is inefficient and unsafe, achieving a First-Try Success Rate nearly 3$\times$ lower than the full system.

It is the synergistic cycle of deduction and abduction that enables success. AGEL-Comp \textit{deduces} plans with logical soundness and \textit{abductively} synthesizes new, generalizable rules (Horn clauses) when its world model proves incomplete. This allows the agent to build an explicit, interpretable, and compositionally structured understanding of its world from the ground up. This work provides a robust technical foundation and a principled path toward agents that can truly learn, reason, and generalize in complex, interactive environments.

\section{Code Availability}

Source code for the AGEL-Comp framework, the Retro Quest simulation environment, and experiment scripts are available on GitHub: \color{blue}\texttt{\url{https://github.com/Place-Beyond-Bytes/AGEL-Comp}}

\color{black}
\section{Declaration on Generative AI}

The authors declare that generative AI tools were used solely for the 
purpose of improving language, grammar, and readability in the manuscript. 
All intellectual contributions, ideas, and arguments 
presented in this paper are entirely the authors' own.






\bibliographystyle{splncs04} 
\bibliography{sample}


\end{document}